\documentclass[letterpaper, 10 pt, conference]{ieeeconf}  

\let\labelindent\relax

\usepackage[]{enumitem}

\makeatletter
\let\MYcaption\@makecaption
\makeatother

\usepackage[font=footnotesize]{subcaption}

\makeatletter
\let\@makecaption\MYcaption
\makeatother\usepackage{graphicx}

\usepackage{mathtools, nccmath}
\usepackage{bm}
\usepackage{multirow}
\newcommand{\subparagraph}{} 
\usepackage{titlesec}
\usepackage{booktabs}

\newcommand{\supp}{\textit{supportive}}
\newcommand{\Supp}{\textit{Supportive}}
\newcommand{\etc}{\textit{etc}. }
\newcommand{\etal}{\textit{et al}. }
\newcommand{\eg}{\textit{e}.\textit{g}.}
\newcommand{\vect}[1]{\boldsymbol{#1}}

\IEEEoverridecommandlockouts                              

\title{\LARGE \bf
Supportive Actions for Manipulation in Human-Robot Coworker Teams
\vspace{-10mm}
}

\author{Shray Bansal$^{1}$, Rhys Newbury$^{2}$, Wesley Chan$^{2}$, Akansel Cosgun$^{2}$, Aimee Allen$^2$, \\
Dana Kuli\'{c}$^{2}$, Tom Drummond$^{2}$ and Charles Isbell$^{1}$
\thanks{$^{1}$ School of Interactive Computing, Georgia Institute
of Technology, Atlanta, GA, 30308, USA.
        {\tt\small sbansal34@gatech.edu}}%
\thanks{$^{2}$ Department of Electrical and Computer Systems Engineering, Monash University, Clayton, VIC, Australia.
        }%
\vspace{-2mm}
}

\begin{document}

\maketitle
\thispagestyle{empty}
\pagestyle{empty}

\begin{abstract}
The increasing presence of robots alongside humans, such as in human-robot teams in manufacturing, gives rise to research questions about the kind of behaviors people prefer in their robot counterparts. We term actions that support interaction by reducing future interference with others as \supp~robot actions and investigate their utility in a co-located manipulation scenario. We compare two robot modes in a shared table pick-and-place task: (1) Task-oriented:  the robot only takes actions to further its own task objective and (2) \Supp: the robot sometimes prefers \supp~actions to task-oriented ones when they reduce future goal-conflicts. Our experiments in simulation, using a simplified human model, reveal that supportive actions reduce the interference between agents, especially in more difficult tasks, but also cause the robot to take longer to complete the task. We implemented these modes on a physical robot in a user study where a human and a robot perform object placement on a shared table. Our results show that a \supp~robot was perceived as a more favorable coworker by the human and also reduced interference with the human in the more difficult of two scenarios. However, it also took longer to complete the task highlighting an interesting trade-off between task-efficiency and human-preference that needs to be considered before designing robot behavior for close-proximity manipulation scenarios. 
\end{abstract}

\section{Introduction}

Despite the continued growth of industrial robot sales~\cite{robot_report}, many assembly tasks are still performed manually in major industries
~\cite{unhelkar2014comparative}. A vision for the future of manufacturing involves robots working alongside human coworkers on tasks that exploit the respective strengths of both. Surveys identify interaction with co-workers as one of the most important job criteria for human workers~\cite{welfare2019consider}. 
We introduce interaction-supporting actions that aim to improve the coworker experience in human-robot co-located manipulation.  
We implement these in a close-proximity manipulation task to understand the impact on task performance and the coworker perception as compared to a robot focused solely on completing its task.

\begin{figure}[ht!]
  \centering
  \includegraphics[trim=2.5cm 6cm 0cm 0.5cm,clip,width=0.95\linewidth]{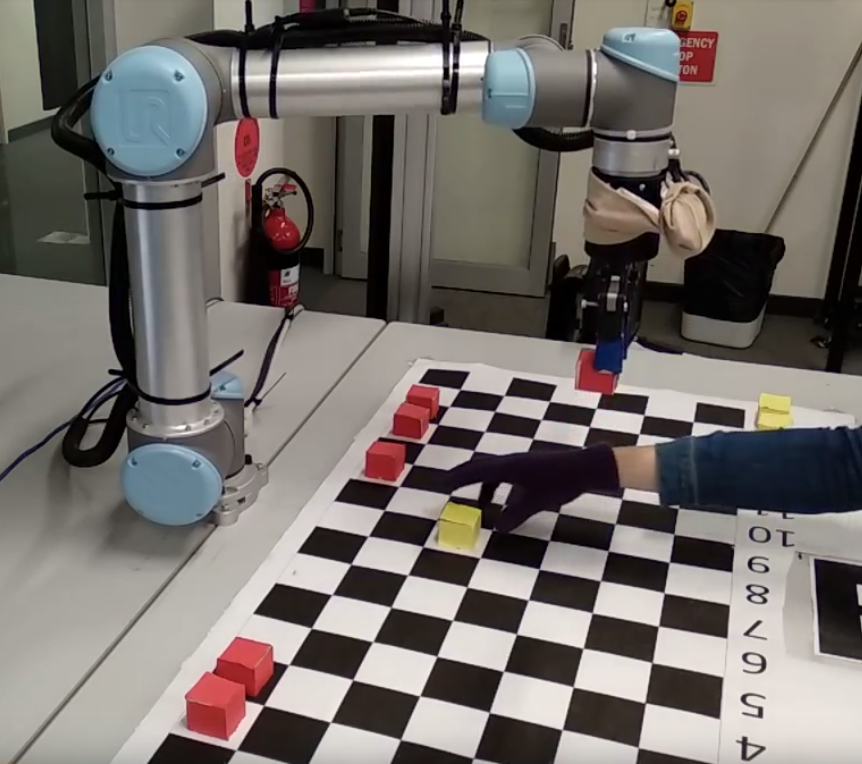}
  \caption{An example scene from our co-located manipulation scenario. The robot's goal is to place all the red blocks into the row closest to itself, and the human participant's goal is to do the same for the yellow blocks.}
\vspace{-8mm}
    \label{fig:picking_together}
\end{figure}

We term actions necessary for an agent to complete their task in the absence of other agents as \textit{task-oriented}. We define \supp~actions as actions that support the interaction by reducing potential interference with other agents but are not necessary for task completion. For \eg, when resetting a chessboard, for the agent playing black, actions that move the black pieces to their positions are \textit{task-oriented} and actions moving the white pieces towards the other player can be \supp. Although the \supp~actions help the other agent, they are not altruistic as the agent hopes to benefit from the reduced interference they cause. Humans also perform \supp~actions, perhaps due to their modeling of others as intentional agents that plan for mutual benefit \cite{stout1999planning, hoffman2007ensemble}, or their expectations of reciprocity~\cite{fehr2001theories}, \etc

Our task is inspired by other close-proximity human-robot interaction (HRI) manipulation studies \cite{gabler2017game, mainprice2016goal}. It involves two agents, a human and a robot, situated across a table scattered with color-coded blocks, each aiming to bring the blocks of their assigned color quickly back to their side (Fig. ~\ref{fig:picking_together}).
The agents have intentionally been assigned separate goals without a direct incentive for cooperation and the shared table is expected to induce interference. We focus on high-level decision-making and design \supp~actions that proactively avoid collision by modifying the goal configuration of the other agent by moving their blocks.
In our experiments, the robot operates in one of two modes: (1) \textit{Task-oriented}, where the robot takes only \textit{task-oriented} actions, and (2) \Supp~mode where the robot takes both \textit{\supp} and \textit{task-oriented} actions depending on the situation. We hypothesize that the \supp~mode would reduce interference and lead to a better human experience, both in terms of subjective and objective measures. We test this in simulation with a simplified human model and verify it with a user study on a physical robot. 

Our main contributions are the introduction of supportive actions in a human-robot collaborative manipulation task, simulation experiments and user study experiments that justify the use of such actions, and the identification of a trade-off between operational and usability metrics when the robot is designed to deliberately take $\supp$ actions.

After reviewing the literature in Sec.~\ref{sec:related_works}, we formulate the problem in Sec.~\ref{sec:problem_formulation} and describe our methodology in Sec.~\ref{sec:method}. We first experiment in simulation in Sec.~\ref{sec:simulation} to design the \supp~actions and formulate hypotheses in Sec.~\ref{sec:hypotheses}. We present the implementation and user study details in Sections~\ref{sec:implementation} and~\ref{sec:user_study}, respectively. We analyze results in Sec.~\ref{sec:results}, discuss them in Sec.~\ref{sec:discussion} and conclude in Sec.~\ref{sec:conclusion}.

\section{Related Work}
\label{sec:related_works}

Human-robot interaction (HRI) includes collaborative scenarios where agents work to achieve a common goal and others where agents have separate, sometimes competing, objectives. Our goal is to study scenarios where humans and robots work alongside each other, and interaction arises from conflict due to shared resources (such as space). 

In manipulation, human-robot co-presence focus on scenarios where the human is either treated as an obstacle to be avoided \cite{mainprice2016goal, li2019safe}, or, as a leader \cite{hawkins2014anticipating} to be assisted. In the former, the human's goal is either not considered at all, or only used to make predictions to guide a more pro-active obstacle avoidance, and in the latter, the robot shares the human's goal. Our task involves separate goals for the two and we consider the question of whether the robot should take actions that support the interaction without direct task completion benefits for the robot. 

For assisting the human by anticipating their actions, Hawkins \etal~\cite{hawkins2014anticipating} exploit task structure and Nikolaidis \etal~\cite{nikolaidis2015efficient} perform online adaptation to user preferences. Similarly to us, both of these approaches focus on the high-level decision-making aspect of the task. 
Cherubini \etal~\cite{cherubini2016collaborative} plan low-level robot actions to successfully reduce human workload for automotive manufacturing and Koppula \etal~\cite{koppula2015anticipating} perform assistive actions adapted to predictions made by learning model of the human's activities.

These methods have helped improve collaborative task performance, but the inherent assumption is that the role assigned to the robot should be to assist and/or stay out of the way. These simplify the robot's decision-making to favor actions that directly further the human's objective. However, the types of roles and interaction modes in mixed human-robot teams are richer, as shown by Gombolay \etal~\cite{gombolay2015decision}. 

Similar to our task, Gabler \etal~\cite{gabler2017game} plan robot actions in a close-proximity human-robot collaborative scenario. Although both agents have a common goal in their task, they utilize a game-theoretic model that considers the human as an agent with a different utility. They use this goal-driven behavior while planning to increase joint task-efficiency. We design \Supp~actions to influence human behavior by modifying goal configuration by moving their blocks. This helps to avoid future interference. While their model also considers the robot's influence on the human, they only use it to find an optimal ordering of existing task-oriented actions.  

\section{Problem Formulation}
\label{sec:problem_formulation}
\begin{figure}
  \centering
  \includegraphics[width=0.6\linewidth]{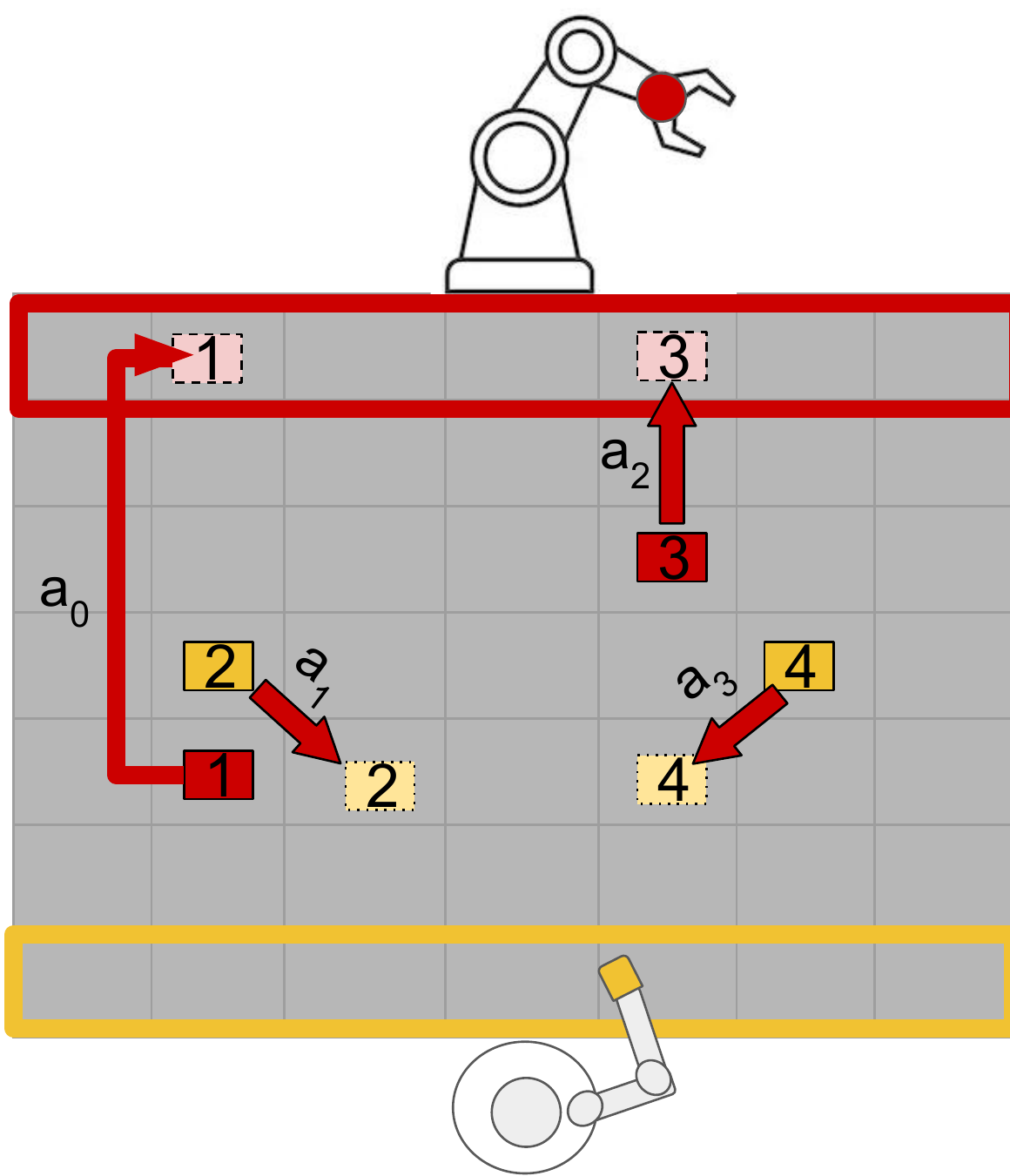}
  \caption{An example board configuration consisting of blocks ($1-4$) for the robot (red) and the human (yellow). Robot actions ($a_{0-3}$) are depicted by the arrows. $a_0, a_2$ are \textit{task-oriented} actions while $a_1, a_3$ are \supp~and $a_1$ is a more useful \textit{\supp} action because it reduces potential interference when reaching for block $1$.}
  \label{fig:example_scenario}

\end{figure}

We design a pair of pick-and-place tasks on a table shared by a human and a robot and represent it as a two-agent game. 
The table has two sets of blocks distinguished by color, we assign one set to the robot $\vect{b_R} = \{b^1_R, ..., b^n_R\}$ and the other to the human $\vect{b_H} = \{b^1_H, ..., b^n_H\}$. We draw a $2D$ grid on the table and place each block in a single cell. We define this cell as the block's location $l(b^i) = (r,c)$ and a cell near its assigned agent as its destination, $d(b^i) = (r,c)$.
A state is a configuration of blocks on the grid, $s = \{b_R^i, b_H^i\} \quad \forall i \quad 1 \leq i \geq n$. 
Fig.\ref{fig:example_scenario} shows a grid configuration where $n=2$ and $\vect{b_R}=\{\bm{1},\bm{3}\}$ and $\vect{b_H}=\{\bm{2},\bm{4}\}$. 

In this task, an action $a$ can move at most one block to a different location. For instance, in Fig. \ref{fig:example_scenario}, $a_0 = (\bm{1}, d(\bm{1}))$, moves block $\bm{1}$ from its location to the goal. We also allow idle actions 
that do not move any blocks. 
Both agents are instructed to start performing actions simultaneously, and so, if one agent finishes their action early, they have to wait for the other to complete their action before starting to perform the next one. 
We assign each agent the goal to take actions that lead to a state, $s^{G}$, where each of their blocks is in its destination cell, in minimum time. Their goal only depends upon their blocks, the locations of other blocks is not  directly relevant.

\section{Method}
\label{sec:method}

We first explain how to construct the sets of \textit{task-oriented} and \supp~actions and then describe two decision-making strategies used by the robot to perform the task.

\subsection{Action Sets}

We define two action sets for the robot to use: task-oriented, $A^{TO}(s)$ and \supp, $A^{S}(s)$. 
$A^{TO}$ includes actions that each move a robot block to its destination,
\begin{equation}
    A^{TO}(s) = \{a = (b_R, d(b_R)) \mid \forall b_R, l(b_R) \neq d(b_R)\}.
\end{equation}

$A^{S}$ includes the {\supp} actions. We define a {\supp} action, $a = (b_H, d)$, for a human block, $b_H$, that is closer to the robot. Then we set the closest empty cell to it, which is also nearer to the human, as its destination $d$. 
This way, we balance the cost of the additional action with reducing the potential for interference while favoring the human's preference of retrieving objects near them. For \eg, in Fig. \ref{fig:example_scenario}, $ A^{TO} = \{a_0, a_2\}$ and $A^{S} = \{a_1, a_3\}$.

\subsection{Task-Oriented Robot}
The \textit{task-oriented} baseline randomly samples an action from the \textit{task-oriented} set at a given state, $a_R \sim A_R^{TO}(s)$. The goal is to complete the task with the fewest actions. It chooses randomly because all \textit{task-oriented} actions are necessary for reaching the  goal state. 

\subsection{Supportive Robot}

The {\Supp} robot chooses actions using a policy, $\pi$ containing actions from \textit{task-oriented} and {\supp} sets.
This policy is an ordered set of actions ranked by their priority and is defined by the user before starting the task. 
Here, we describe the heuristical approach we took to create $\pi$ for the task with the goal to reflect the utility of \textit{\supp} actions.

We initialize $\pi$ as an empty list and populate it by iterating over the following rules until no new action is generated. We also initialize \texttt{B} to a list of all the blocks in the grid. 
\begin{enumerate}
    \item Return empty if \texttt{B} is empty.
    \item If a block $b_R^i \in \texttt{B}$ exists such that $b_R^i$ has no human block that might cause a conflict when reaching for it, then pop $b_R^i$ and return a task-oriented action for it.
    \item Else, find a \supp~action from \texttt{B} that has conflict with the most robot objects in \texttt{B}.  
\end{enumerate}
This approach is designed to produce actions that reduce the probability of collision between the human and the robot while trying the minimize the task completion time. It is applicable to any block configuration.  

Given a predefined policy, $\pi$, the robot checks the list in order and executes the first action that is feasible in the current state $s$. If no feasible action is found, it defaults to sampling available \textit{task-oriented} actions until the goal is reached.
We assign a fixed list to $\pi$ to ensure that the participants observe similar behavior from the robot every trial when studying the effect of {\supp} actions.

Fig.\ref{fig:example_scenario} depicts an example task with four blocks, \textit{task-oriented} actions, $A^{TO} =\{a_0, a_2\}$, and \supp~actions, $A^{S} = \{a_1, a_3\}$. The policy, $\pi$, for this scenario is $\pi = (a_2, a_1, a_0)$. Here, a \textit{task-oriented} action, $a_2$, is included first because of the lack of potential goal conflict of block $\bm{3}$; then the robot takes a \textit{\supp} action, $a_1$, to reduce the potential interference of block $\bm{2}$; and finally, it completes the task with the last \textit{task-oriented} action $a_0$. The planner ignores \supp~action, $a_3$ because block $\bm{4}$ causes no potential interference with the robot's blocks.

\section{Simulated Experiment}
\label{sec:simulation}

\begin{figure}
  \centering
  \includegraphics[width=0.5\linewidth]{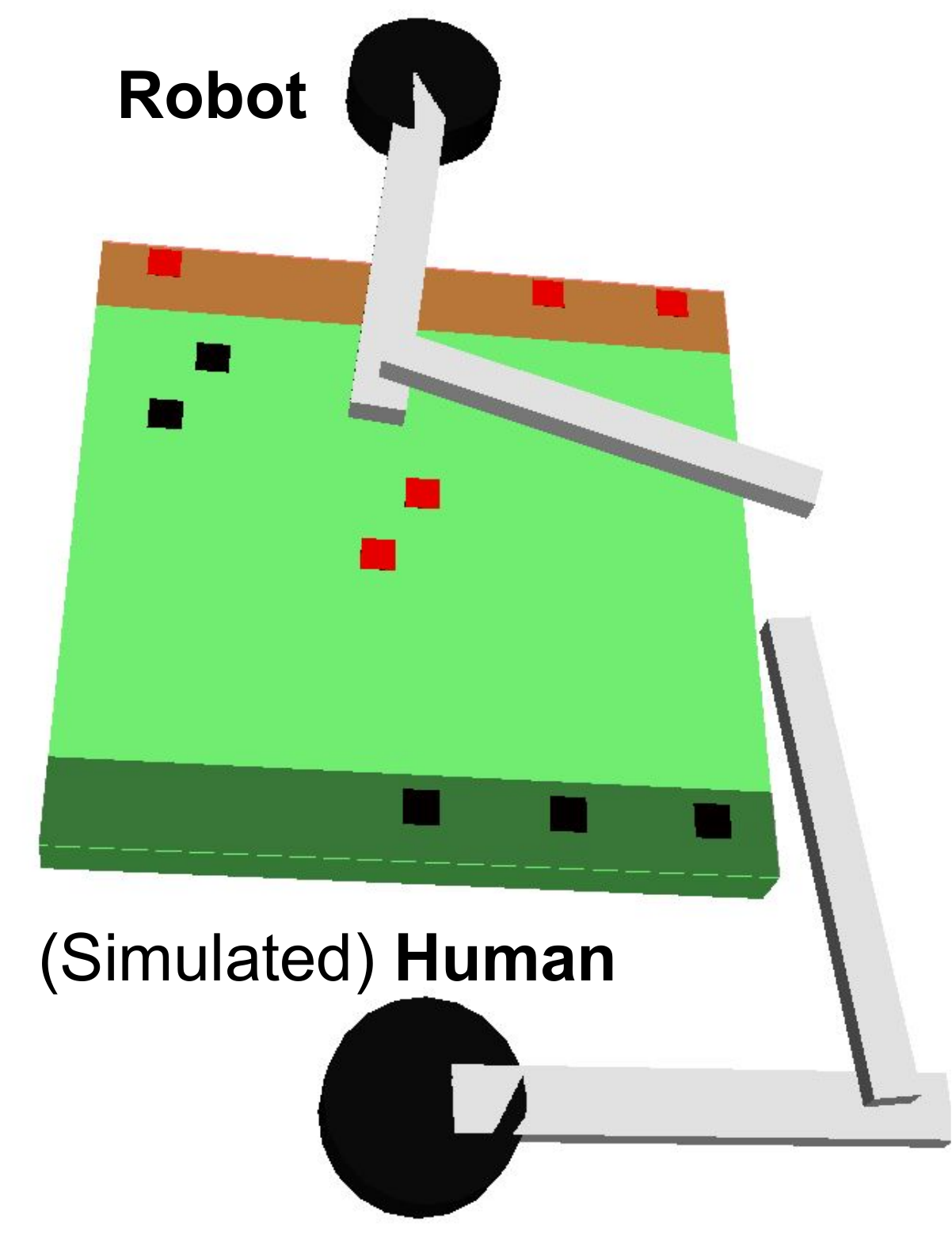}
  \caption{The simulated 2D environment with two arms, one is a simulation of the human and the other is controlled by the robot policy.}
  \label{fig:simulation_scenario}
\end{figure}

We simulate a scenario with two $2$-link robot arms performing pick-and-place actions in $2$D (Fig.~\ref{fig:simulation_scenario}). Our goal is to observe the effect of \textit{\supp} actions in an idealized setting, without the variance introduced by the participants, or errors in sensing and actuation.

\begin{figure*}[ht!]
  \centering
  \begin{tabular}[b]{c}
    \includegraphics[trim=0.25cm 0.5cm 0.25cm 0.5cm, clip, width=.4\linewidth, angle=180]{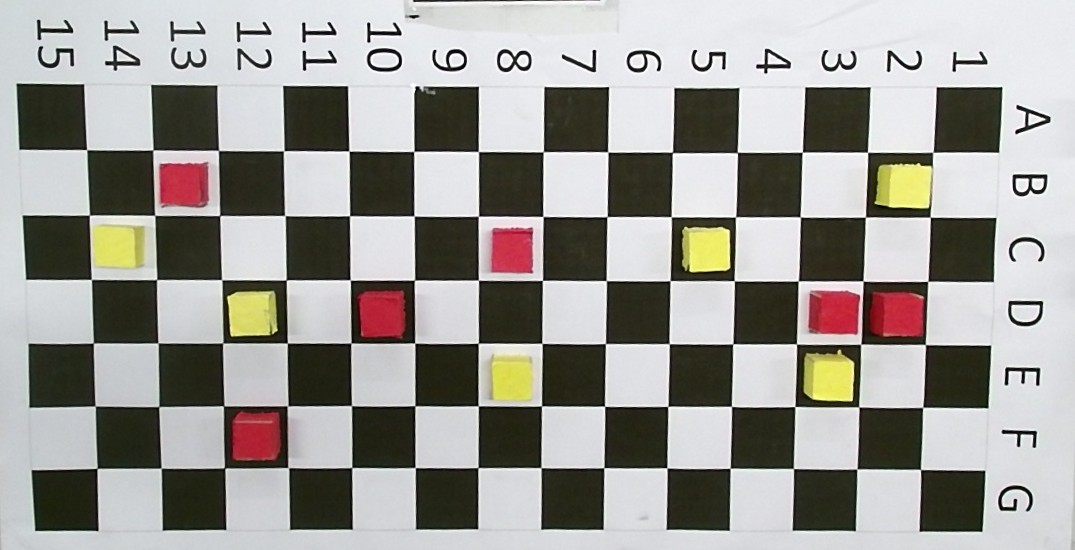} \\
    \small (a) Easy
  \end{tabular} 
  \begin{tabular}[b]{c}
    \includegraphics[trim=0.25cm 0.25cm 0.25cm 0.5cm, clip, width=.4\linewidth, angle=180]{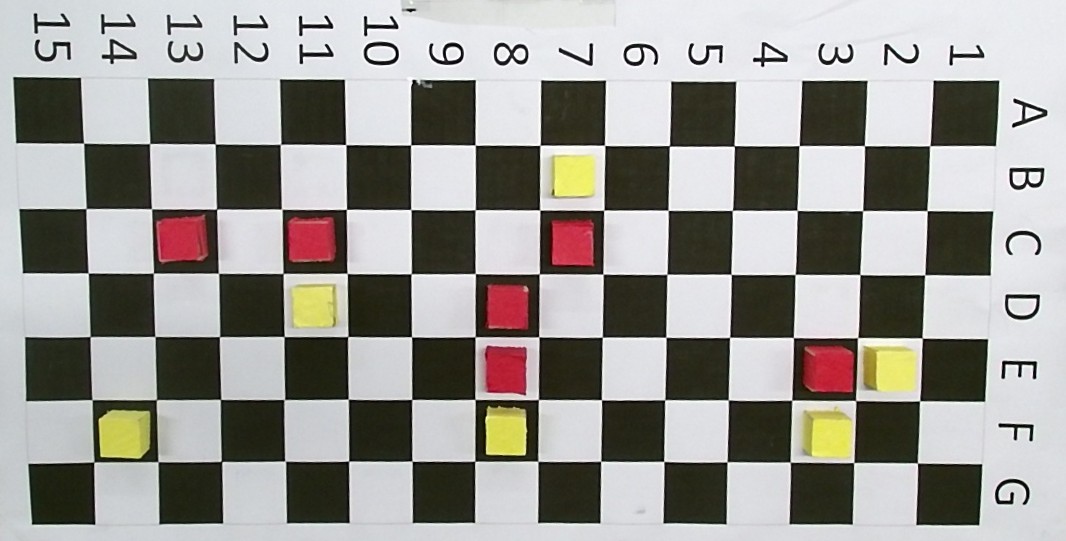}\\
    \small (b) Hard
  \end{tabular}
  \caption{Layout of the \textit{easy} (left) and \textit{hard} (right) block configurations, viewed so the human is seated below row A. The human places yellow blocks on the numbers below row A, whereas the robot is across the table and placing red blocks in Row G. The difficulty is due to the conflict caused by the robot and human reaching for the same space. This conflict exists more in (b) since most of the yellow blocks are in front of the robot's.}
  \label{fig:board_layouts}
\end{figure*}

We develop an OpenRAVE \cite{diankov_thesis} environment with blocks of two colors scattered on a table. We assign each arm six blocks of the same color. The goal for each arm was to bring blocks of their assigned color to the destination area near the arm, highlighted in Fig. \ref{fig:simulation_scenario}. We define a $7\times15$ grid on the table and place the blocks into these cells according to two configurations, \textit{easy} and \textit{hard}, as shown in Fig. \ref{fig:board_layouts}. We consider one arm as the robot and the other one as a simulated human. The simulated human chooses \textit{task-oriented} actions while prioritizing closer blocks. We experiment with the robot following both task-oriented and \supp~algorithms from Sec. \ref{sec:method}. The RRT* \cite{lavalle1998rapidly} implementation in OMPL~\cite{sucan2012the-open-motion-planning-library} is used to plan joint-space trajectories.

\begin{table}[]
\caption{Simulation Results}
\label{tab:sim_results}
\centering
\begin{tabular}{|l|l|l|l|}
\toprule
\textbf{Scenario} & \textbf{Robot Mode} & \textbf{Task Time (s)} & \textbf{Safety Stops} \\
\midrule

\multirow{2}{*}{\textit{Easy}} & \textit{Task-Oriented} & $15.46\pm 0.3$ & $4.6 \pm 0.7$\\
 & \Supp & $17.75\pm 0.2$ & $3.0 \pm 0.7$\\

\midrule

\multirow{2}{*}{\textit{Hard}} & \textit{Task-Oriented} & $15.9 \pm 0.9$ & $7.3 \pm 2.2$\\
 & \Supp & $18.56 \pm 0.5$& $2.4 \pm 1.2$\\
\bottomrule
\end{tabular}
\end{table}

\textbf{Results.} The two scenarios and two robot modes make four experimental conditions. We run each of them $10$ times and present the averaged results in Tab. \ref{tab:sim_results}. The time taken to complete the task by the slowest agent is termed Task Time. We also record the number of times the simulated robot was stopped during the interaction to prevent a collision and term it Safety Stops. The robot stops and waits for the simulated human to move a threshold distance away when this happens while the human is free to move. We find task completion time to be higher for the {\supp} robot but the safety stops are lower in Tab.~\ref{tab:sim_results}. A larger effect due to \supp~actions is observed for both metrics in the \textit{hard} scenario. The \supp~robot is always slower than the human and although the additional actions cause a longer task time they also reduced goal conflict leading to less than $50\%$ safety stops in the \textit{hard} scenario.

\pagebreak
\section{Hypotheses}
\label{sec:hypotheses}

Following simulation results, we anticipate the robot's behavior and the initial block configuration to affect collaborative performance. We formulate the following hypotheses to test on a user study with a physical robot. 

\begin{enumerate}[label=\textbf{H\arabic*}]
    \item \textit{\Supp~actions will reduce the interference between the agents.} In particular, we expect the \supp~actions to reduce the safety stops occurring in the interaction, especially for difficult scenarios.
    \item \textit{\Supp~actions will reduce the human's time to complete the task.} We expect people to complete the task faster when interacting with the \supp~robot leading to more idle time, especially for difficult scenarios.
    \item \textit{\Supp~actions will have a positive effect on the subjective measures of task performance.} We expect that participants will prefer the \supp~robot as a coworker, especially for difficult scenarios.
    \item \textit{Changing the initial block configuration would affect both the subjective and objective measures.} In particular, we expect that participants will find the task more difficult to perform if the initial block configuration includes more goal conflicts. We also expect the effect of \supp~actions to be more prevalent in difficult scenarios, in general.  
\end{enumerate}

\section{User Study Design}
\label{sec:user_study}

We conduct a user study to test the effect of the \supp~actions. The study was approved by Monash University's Ethics Review Board.

\subsection{Independent Variables} 
We manipulate two independent variables. 
\begin{itemize}
    \item \textbf{Robot mode}: \{\textit{Task-Oriented}, \textit{\Supp}\} robots as described in Section \ref{sec:method}.
    \item \textbf{Scenario}: \{\textit{Easy}, \textit{Hard}\} block configurations. (Fig. \ref{fig:board_layouts})
\end{itemize}

The block configuration in the \textit{easy} and \textit{hard} scenarios are designed to cause different levels of goal conflict. While both of them include six blocks, the robot's blocks in the \textit{hard} scenario were arranged to be directly in front of the human's. 
We expect this would increase task difficulty by causing more interference since both agents need to reach into the same space. 

\subsection{Participant Allocation}
We recruited $18$ subjects aged $20-31$ ($M=22.4$, $SD=3.1$, $11$ male, $7$ female) for a within-subject study. 
To reduce order effects, we counterbalanced the order of the robot mode. We kept the scenario order the same, where \textit{hard} always followed \textit{easy}. The participants were not informed about the kind of robot they would be interacting with or how many types there were. 

\subsection{Procedure}
The experiment took place in a university lab under experimenter supervision. We seated participants in front of the robot as depicted in Fig.\ref{fig:setup}. After reading the explanatory statement and signing a consent form, the experimenter explained the task by reading from a script. 

The participants were assigned yellow blocks and their goal was to move these blocks to their destinations accurately while minimizing task time. The start of a turn was signaled on the scanning display in Fig. \ref{fig:setup} and both agents performed reaching actions simultaneously, continuing until all their blocks were in their respective destinations. This concluded one trial and each participant performed four.   
Participants were also given three types of surveys, a demographic one at the start of the experiment, one after every trial, and one at the end to record their overall experience. A complete experiment took between $30$ and $45$ minutes.

\subsection{Dependent Variables}
We record both objective and subjective metrics.

\noindent\textbf{Objective measures. } We study the effect of {\supp} actions on task completion time for each agent, the total number of safety stops, as well as human's idle time ratio. 
The task completion time is the time an agent takes to complete a trial and is easily measured for the robot since we programmatically record the time when the robot starts and finishes an action. For the human, we manually annotate this using a video recording of the experiments. We also annotate the time the human waits for the robot after completing an action and compute the ratio of the accumulated wait time over a trial to their total execution time as the human-idle ratio. 
We also count the times the robot has to stop due to proximity to the human as safety stops.

\noindent\textbf{Subjective measures.} Participants answered ten $5$-point questions after each trial. Five of these are collected in a Likert-scale that measures robot proficiency as a coworker and includes statements about the robot's helpfulness, action-selection, intention-prediction, disruption, \etc. The rest of the questions are treated as individual differential scale items. We adapt this survey from collaborative HRI studies like~\cite{hoffman2019evaluating}. The Likert-scale (Cronbach's $\alpha = 0.807$) is listed in Tab.~\ref{tab:questions} and the individual items are listed in Tab.~\ref{tab:individual_questions}. 

\begin{table}
  \caption{Likert-scale composed of individual survey items with Cronbach's $\alpha$. (R) indicates a reverse scale.}
  \label{tab:questions}
  
  \begin{tabular}{|l|}
    \toprule
     \textbf{Robot coworker proficiency ($\alpha = 0.807$)} \\
     I believe the robot accurately perceived my goals.\\
     The robot was helpful and/or cooperative.\\
     The robot seemed to select the correct object to pick up \\ 
     most of the time.\\
     The robot disrupted me in efficiently performing the task. \textbf{(R)} \\
     I felt uncomfortable with the robot. \textbf{(R)} \\
  \bottomrule
\end{tabular}
\end{table}

\begin{table}
\caption{Individual scale items from survey.}
  \label{tab:individual_questions}
  \begin{tabular}{|l|}
  \toprule
\textbf{Individual Measures}\\
    \textbf{I1} How successful were you in achieving your task?\\
    \textbf{I2} How hard did you have to work to accomplish your\\ level of performance? \textbf{(R)}\\
    \textbf{I3} How much attention did you pay to the robot and \\its performance during the task? \\
    \textbf{I4} I felt unsafe with the robot.\textbf{(R)} \\
    \textbf{I5} How would you grade the robot as a coworker, overall?\\
    \bottomrule
    \end{tabular}
\end{table}

\section{Implementation Details}
\label{sec:implementation}

\begin{figure}[]
  \centering
  \includegraphics[trim=2.5cm 1cm 1.2cm 0.4cm,clip, width=1.0\linewidth]{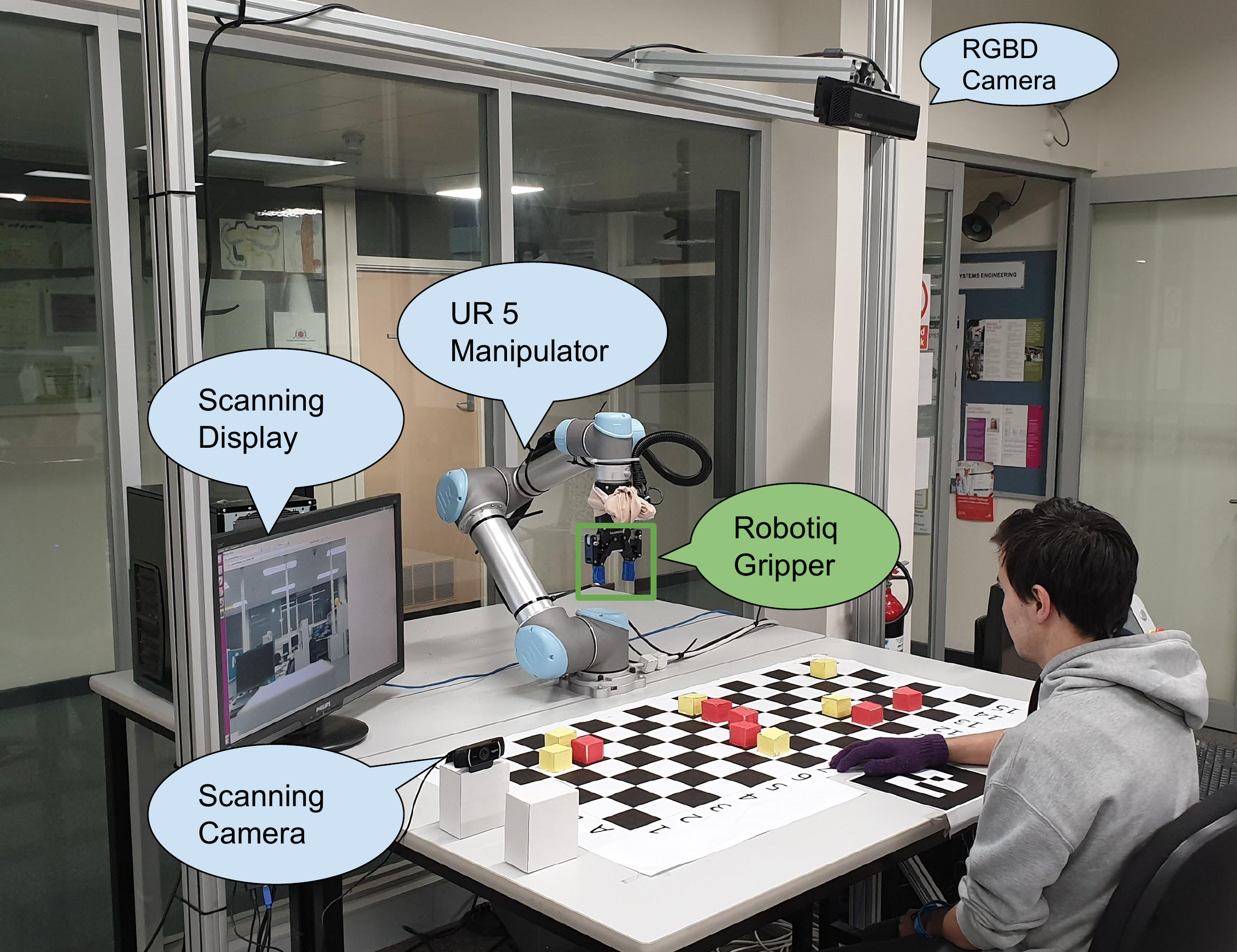}
  \caption{The experimental setup}
\label{fig:setup}
\end{figure}

Our user study setup is depicted in Fig.~\ref{fig:setup} and includes the robot and the human around a table with a checkerboard grid on which we place the blocks. We mount an RGB-D sensor overhead to detect the blocks and the person's arm. 
These detections guide the robot's action-selection and trajectory planning, which are implemented on the Universal Robot 5 (UR5) using the Robot Operating System (ROS)~\cite{ros}.  
We also include a scanning area that instructs the participant about the destinations for their blocks. 
Our experiment is fully-autonomous and does not require human intervention.

\subsection{Sensing}
\label{sec:block_detect}

The location of the grid is calibrated in the camera frame ahead of time using OpenCV~\cite{OpenCV} and we apply a simple color blob detection technique to the RGB image in real-time to localize the blocks.

We instructed participants to wear a colored glove covering their arm to allow for its easy detection. We ensure safety by stopping the robot arm if the user's hand comes within a fixed distance threshold.

\subsection{Robot Control}
\label{sec:robot_control}

We implement both \textit{task-oriented} and \supp~robot policies for action-selection. For a given goal grid location, we generate waypoints for the robot end-effector to it at a fixed vertical offset from the grid and use the MoveIt framework~\cite{moveit} to generate a Cartesian path. This path is followed by the robot controller after which it attempts a vertical move down to either grab or drop the block and then moves back up. Robot joint speed is limited to ($0.314 rad/s$) to ensure user safety and comfort. 

We also included a camera station where participants scanned blocks and were informed of their destinations after a short delay. We use this delay to account for the human's higher relative speed to synchronize human-robot actions.

\section{user study Results}
\label{sec:results}

We compare the independent variables through the objective task performance metrics first and then by participant responses to the survey. We had to remove the data for two participants, one due to a robot failure, and the other because the participant did not follow experimental directions. Thus, in total we analyze ($N = 16)\times4 = 64$ trials. 

\subsection{Objective Measures}

\begin{figure*}[t]
  \centering
  \begin{subfigure}{.32\textwidth}
    \includegraphics[trim=0.9cm 0.cm 0.cm 0.cm, clip, width=.9\linewidth]{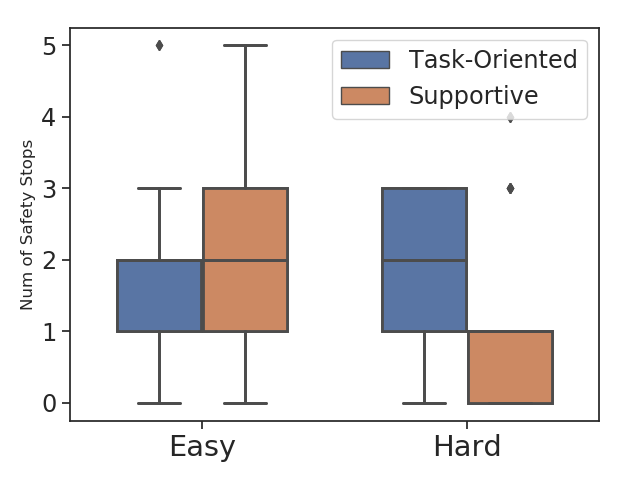}
    \caption{Safety Stops}
    \label{fig:safety_ob}
  \end{subfigure} 
  \begin{subfigure}{.32\textwidth}
    \includegraphics[trim=0.2cm 0.5cm 0.cm 0.cm, clip, width=.95\linewidth]{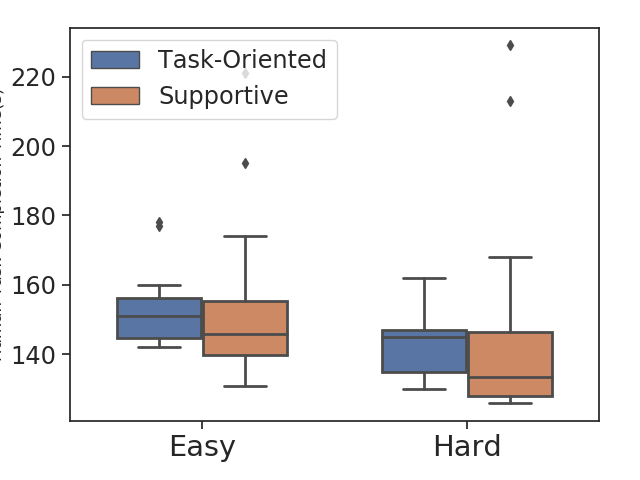}
    \caption{Human Task Time}
    \label{fig:htime_ob}
  \end{subfigure}
  \begin{subfigure}{.32\textwidth}
    \includegraphics[trim=0.5cm 0.5cm 0.cm 0.cm, clip, width=.9\linewidth]{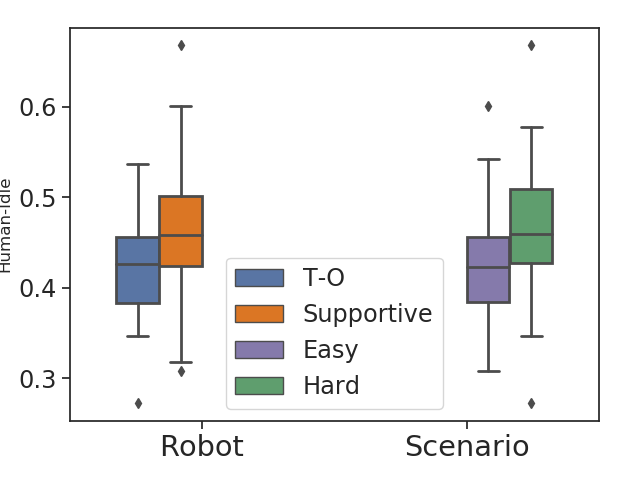}
        \caption{Human Idle Time}
    \label{fig:h_idle_ob}
  \end{subfigure}
  \caption{Objective Measures. Box-and-whisker plots of the (a) number of safety stops; (b) time taken by the human to complete the task; and (c) the proportion of idle time spent by the human. Note, T-O refers to the \textit{task-oriented} robot.}
  \label{fig:objective_measures}
\end{figure*}

We analyze some of the objective metrics in Fig. \ref{fig:objective_measures}. 

\textbf{Safety Stops.} We count the times when the robot has to stop due to its proximity to the human's arm. We compare robot types through a Wilcoxon signed-rank test on each scenario because the data was not normally distributed. We find a significant effect due to the \supp~robot in the \textit{hard} scenario ($w=79.5, p<0.05$). Fig. \ref{fig:safety_ob} shows that the \supp~robot had fewer stops in \textit{hard}  affirming \textbf{H2}.

\textbf{Robot Task Time.} We use a repeated-measure two-way ANOVA to compare the robot's task completion time. We find a significant effect due to the \supp~robot ($F(3,60)=74.0, p<0.01$) and no interaction. Table \ref{tab:robot_time} shows that the addition of \supp~actions led to a longer robot task time.
\begin{table}[h]
\vspace{-2mm}
\centering
\caption{Task completion time of the robot.}
\label{tab:robot_time}
\begin{tabular}{l|l}
\toprule
Robot & Robot Task Time (s)\\
\hline
Baseline & $162.9 \pm 3.9$ \\
\Supp & $208.9 \pm 3.7$ \\
\bottomrule
\end{tabular}

\vspace{-3mm}
\end{table}

\textbf{Human Task Time.} We use a Wilcoxon signed-rank test to compare the human's task completion time due to the non-normality of this data. We find no significant effect due to \supp~actions for either scenario. Fig. \ref{fig:htime_ob} shows the human interacting with the \supp~robot is faster but with high variance, partly denying \textbf{H3}.

\textbf{Human-Idle Time.} We use a repeated-measures two-way ANOVA to analyze the human's idle time ratio as a measure of task fluency. We compute this ratio by accumulating the time the human waited for the robot to complete an action before they could start the next one and dividing it by the human's task time. We find significant effects due to both robot ($F(3,60)=7.3, p<0.05$) and scenario ($F(3,60)=5.95, p<0.05$) types. Fig. \ref{fig:h_idle_ob} shows that \supp~robot and \textit{hard} scenario each led to higher idle time partially affirming \textbf{H3}. This measure was adapted from~\cite{hoffman2019evaluating} where it was found to be correlated with higher human preference. 

\textbf{\Supp~actions.} The robot took on average fewer \supp~actions in the \textit{easy} ($1.9$) scenario than the the \textit{hard} ($2.6$) due to fewer goal conflicts. The participants took only $5$ \supp~actions overall and all of them took place in the \supp~robot condition.

\textbf{Summary.} The \supp~robot confirms \textbf{H1} in the \textit{hard} scenario by reducing interference; it partly confirms \textbf{H3} since human's idle time is increased, however, the human's task completion time is not significantly reduced. Also, the \supp~robot takes longer to complete this task.

\subsection{Subjective Measures}

\begin{figure*}
  \centering
  \begin{subfigure}{.32\textwidth}
    \includegraphics[trim=0.cm 0.cm 0.cm 0.cm, clip, width=.9\linewidth]{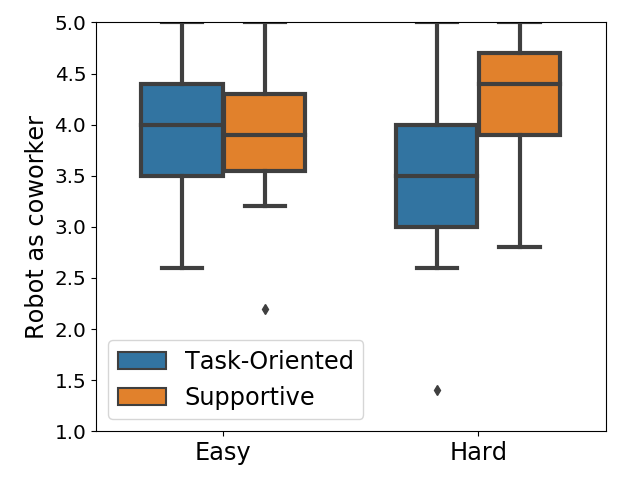}
    \caption{Likert Scale}
    \label{fig:likert_sub}
  \end{subfigure} 
  \begin{subfigure}{.32\textwidth}
    \includegraphics[trim=0.cm 0.cm 0.cm 0.cm, clip, width=.9\linewidth]{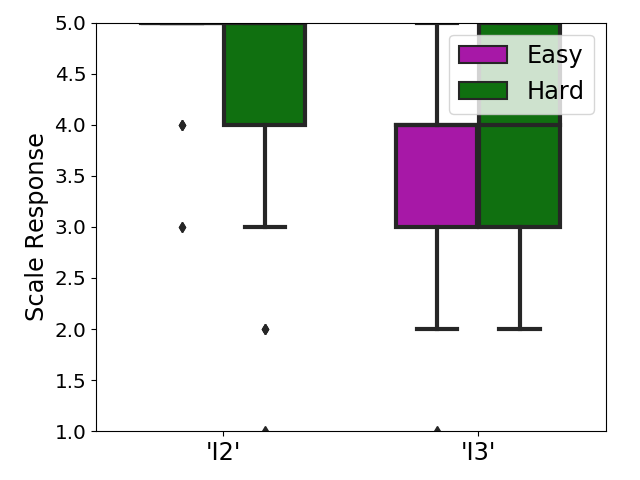}
    \caption{Individual Measures}
    \label{fig:individual_sub}
  \end{subfigure}
  \begin{subfigure}{.32\textwidth}
    \includegraphics[trim=0.cm 0.cm 0.cm 0.cm, clip, width=.9\linewidth]{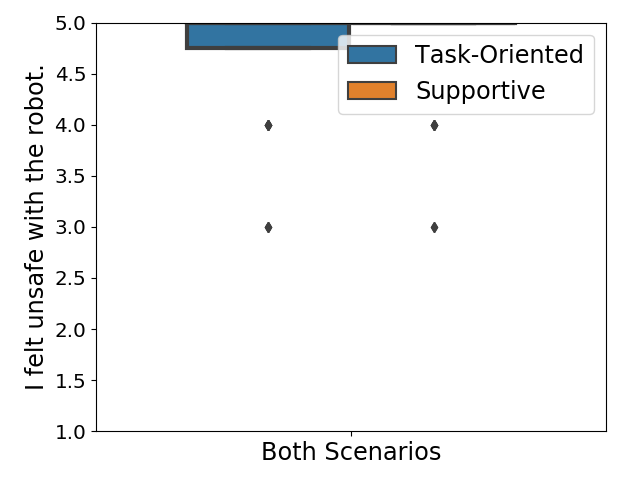}
        \caption{Safety Perception}
    \label{fig:safety_sub}
  \end{subfigure}
  \caption{Subjective Measures. Box-and-whisker plots of the (a) normalized survey response to Likert-scale items for the different robot type separated by scenario; (b) response to measures of subjective task difficulty and attention to the robot for the two scenarios; and (c) safety perception for the robot types. Note that the leftmost box in (b) and rightmost box in (c) have no height and so appear as a line at $5.0$.}
  \label{fig:subjective_measures}
  \vspace{-3mm}
\end{figure*}

We analyze some of the survey responses
in Fig. \ref{fig:subjective_measures}.

\textbf{Robot coworker proficiency.} We perform a two-way repeated-measure ANOVA on the Likert-scale from Table \ref{tab:questions} and find significant interaction ($F(3,60)=13.9$, $p<0.01$). The normalized responses in Fig. \ref{fig:likert_sub} show that participants prefer the \supp~robot in the \textit{hard} scenario but have no preference in the \textit{easy} one affirming \textbf{H1} for it. 
They also show that people prefer \supp~robot more when the task difficulty increases but preference for the \textit{task-oriented} robot remains similar regardless of task difficulty.

\textbf{Scenario Effect}. We use a Wilcoxon signed-rank test to compare individual scale responses from Table \ref{tab:individual_questions}. We find significant scenario effect for both \textbf{I2} ($w=0.0$, $p<0.01$) and \textbf{I3} ($w=34.0$, $p<0.05$). Fig. \ref{fig:individual_sub} indicates that participants find the \textit{hard} scenario more difficult to perform, affirming \textbf{H4}. It also leads to the observation that people are more observant of the robot's actions in the \textit{hard} scenario.

\textbf{Safety Perception}. We used a Wilcoxon signed-rank test to compare the \textbf{I4} scale item and do not find any significant effect due to \supp~actions. Fig. \ref{fig:safety_sub} shows that participants felt very safe for both robot types in our experiment.

\textbf{Summary}. We find that participants prefer the \supp~robot as their coworker in the \textit{hard} scenario affirming \textbf{H3}; also, participants find the \textit{hard} scenario more difficult and pay more attention to the robot in it, supporting  \textbf{H4}.

\section{Discussion} 
\label{sec:discussion}
One might think that moving the human's blocks close to them would cause people to perceive the robot as helpful and inflate \supp~robot's proficiency. However, our results, which show that the \supp~robot is only preferred in the \textit{hard} scenario, provide evidence for the human's preference relying on the suitability of the robot's action-selection to the task. 

Our results show that \supp~actions do not reduce safety stops in the \textit{easy} scenario. Safety stops are overlaps in agent trajectories and can be caused by an unavoidable conflict between agent goals, uncertainty about each other's goal, sensor error, etc. We label a configuration as \textit{hard} due to the presence of more goal conflicts; this label does not allude to other sources of overlap. Supportive actions in our work were designed to reduce goal conflicts, they lead to fewer stops on the \textit{hard} task, but will need to be adapted for other sources of conflict to be effective in other scenarios. 

Hoffman~\cite{hoffman2019evaluating} found that collaborative fluency does not track task-efficiency in team tasks. Ours is not a team task, however, our results also show coworker acceptance to be separate from either agent's task-efficiency. We find \supp~actions to increase coworker acceptance but reduce robot efficiency. They present a trade-off that needs to be considered for designing robot behaviors. For \eg, if a robot is introduced into a manual process to reduce repetitive tasks for humans and increase their job satisfaction, then its acceptance might play a more important role than its efficiency. Our methodology helps highlight this trade-off by combining the subjective and objective impact of supportive robot behaviors and is applicable to other shared-workspace human-robot environments. We consider this methodology as one of the contributions of our work.    

\section{Conclusion and Future Work}
\label{sec:conclusion}

We introduce interaction-supporting actions and design robot behavior that selects between these and task-oriented actions by considering the human's and its own goals. We implement it on an autonomous robot and evaluate it in a shared-workspace user study. The results show that this robot increases human coworker preference in a scenario with more goal conflicts but decreases efficiency as compared to a robot that only takes task-oriented actions. 

Our study illustrates taking actions to support interaction while trading off on efficiency in an assembly task. Although, the rationale from Sec. \ref{sec:method} can help guide adaptation to new domains, however, the actions are applicable only to similar scenarios. In future work, we plan to develop a framework for \supp~behavior that can perform this reasoning based on task-specific cost functions.

Participants took very few \supp~actions towards the robot. We believe their unfamiliarity with the task caused uncertainty about allowed actions. An interesting extension would be to apply this to an actual manufacturing task with subjects who are familiar with it to test the generalizability of our findings. We can also improve task naturalness by increasing robot speed by employing better sensors and models for human motion prediction.

\bibliographystyle{IEEEtran}
\bibliography{refs}

\begin{thebibliography}{10}
\providecommand{\url}[1]{#1}
\csname url@samestyle\endcsname
\providecommand{\newblock}{\relax}
\providecommand{\bibinfo}[2]{#2}
\providecommand{\BIBentrySTDinterwordspacing}{\spaceskip=0pt\relax}
\providecommand{\BIBentryALTinterwordstretchfactor}{4}
\providecommand{\BIBentryALTinterwordspacing}{\spaceskip=\fontdimen2\font plus
\BIBentryALTinterwordstretchfactor\fontdimen3\font minus
  \fontdimen4\font\relax}
\providecommand{\BIBforeignlanguage}[2]{{%
\expandafter\ifx\csname l@#1\endcsname\relax
\typeout{** WARNING: IEEEtran.bst: No hyphenation pattern has been}%
\typeout{** loaded for the language `#1'. Using the pattern for}%
\typeout{** the default language instead.}%
\else
\language=\csname l@#1\endcsname
\fi
#2}}
\providecommand{\BIBdecl}{\relax}
\BIBdecl

\bibitem{robot_report}
{International Federation of Robotics (IFR)}, ``Ifr press release,'' 2019,
  \url{https://ifr.org/ifr-press-releases/news/robot-investment-reaches-record-16.5-billion-usd},
  Last accessed on 2019-10-01.

\bibitem{unhelkar2014comparative}
V.~V. Unhelkar, H.~C. Siu, and J.~A. Shah, ``Comparative performance of human
  and mobile robotic assistants in collaborative fetch-and-deliver tasks,'' in
  \emph{ACM/IEEE International Conference on Human-Robot Interaction (HRI)},
  2014.

\bibitem{welfare2019consider}
K.~S. Welfare, M.~R. Hallowell, J.~A. Shah, and L.~D. Riek, ``Consider the
  human work experience when integrating robotics in the workplace,'' in
  \emph{2019 ACM/IEEE International Conference on Human-Robot Interaction
  (HRI)}, 2019.

\bibitem{stout1999planning}
R.~J. Stout, J.~A. Cannon-Bowers, E.~Salas, and D.~M. Milanovich, ``Planning,
  shared mental models, and coordinated performance: An empirical link is
  established,'' \emph{Human Factors}, vol.~41, no.~1, pp. 61--71, 1999.

\bibitem{hoffman2007ensemble}
G.~Hoffman, ``Ensemble: fluency and embodiment for robots acting with humans,''
  Ph.D. dissertation, Massachusetts Institute of Technology, 2007.

\bibitem{fehr2001theories}
E.~Fehr and K.~M. Schmidt, ``Theories of fairness and reciprocity-evidence and
  economic applications,'' 2001.

\bibitem{gabler2017game}
V.~Gabler, T.~Stahl, G.~Huber, O.~Oguz, and D.~Wollherr, ``A game-theoretic
  approach for adaptive action selection in close proximity
  human-robot-collaboration,'' in \emph{2017 IEEE International Conference on
  Robotics and Automation (ICRA)}, 2017.

\bibitem{mainprice2016goal}
J.~Mainprice, R.~Hayne, and D.~Berenson, ``Goal set inverse optimal control and
  iterative replanning for predicting human reaching motions in shared
  workspaces,'' \emph{IEEE Transactions on Robotics}, vol.~32, no.~4, pp.
  897--908, 2016.

\bibitem{li2019safe}
S.~Li and J.~A. Shah, ``Safe and efficient high dimensional motion planning in
  space-time with time parameterized prediction,'' in \emph{2019 International
  Conference on Robotics and Automation (ICRA)}, 2019.

\bibitem{hawkins2014anticipating}
K.~P. Hawkins, S.~Bansal, N.~N. Vo, and A.~F. Bobick, ``Anticipating human
  actions for collaboration in the presence of task and sensor uncertainty,''
  in \emph{2014 ieee international conference on Robotics and automation
  (ICRA)}, 2014.

\bibitem{nikolaidis2015efficient}
S.~Nikolaidis, R.~Ramakrishnan, K.~Gu, and J.~Shah, ``Efficient model learning
  from joint-action demonstrations for human-robot collaborative tasks,'' in
  \emph{ACM/IEEE international conference on human-robot interaction}, 2015.

\bibitem{cherubini2016collaborative}
A.~Cherubini, R.~Passama, A.~Crosnier, A.~Lasnier, and P.~Fraisse,
  ``Collaborative manufacturing with physical human--robot interaction,''
  \emph{Robotics and Computer-Integrated Manufacturing}, vol.~40, pp. 1--13,
  2016.

\bibitem{koppula2015anticipating}
H.~S. Koppula and A.~Saxena, ``Anticipating human activities using object
  affordances for reactive robotic response,'' \emph{IEEE transactions on
  pattern analysis and machine intelligence}, vol.~38, no.~1, pp. 14--29, 2015.

\bibitem{gombolay2015decision}
M.~C. Gombolay, R.~A. Gutierrez, S.~G. Clarke, G.~F. Sturla, and J.~A. Shah,
  ``Decision-making authority, team efficiency and human worker satisfaction in
  mixed human--robot teams,'' \emph{Autonomous Robots}, vol.~39, no.~3, pp.
  293--312, 2015.

\bibitem{diankov_thesis}
R.~Diankov, ``Automated construction of robotic manipulation programs,'' Ph.D.
  dissertation, Carnegie Mellon University, Robotics Institute, August 2010.

\bibitem{lavalle1998rapidly}
S.~M. LaValle, ``Rapidly-exploring random trees: A new tool for path
  planning,'' 1998.

\bibitem{sucan2012the-open-motion-planning-library}
I.~A. {\c{S}}ucan, M.~Moll, and L.~E. Kavraki, ``The {O}pen {M}otion {P}lanning
  {L}ibrary,'' \emph{{IEEE} Robotics \& Automation Magazine}, 2012.

\bibitem{hoffman2019evaluating}
G.~Hoffman, ``Evaluating fluency in human--robot collaboration,'' \emph{IEEE
  Transactions on Human-Machine Systems}, vol.~49, no.~3, pp. 209--218, 2019.

\bibitem{ros}
M.~Quigley, K.~Conley, B.~Gerkey, J.~Faust, T.~Foote, J.~Leibs, R.~Wheeler, and
  A.~Ng, ``Ros: an open-source robot operating system,'' 2009.

\bibitem{OpenCV}
G.~Bradski and A.~Kaehler, \emph{Learning OpenCV: Computer vision with the
  OpenCV library}, 2008.

\bibitem{moveit}
\BIBentryALTinterwordspacing
I.~Sucan and S.~Chitta, ``Moveit motion planning framework,'' 2019. [Online].
  Available: \url{http://moveit.ros.org}
\BIBentrySTDinterwordspacing

\end{thebibliography}

\end{document}